\PassOptionsToPackage{numbers,sort&compress}{natbib}

\documentclass[dvipsnames]{article}

\usepackage[final]{openbmb_2025}

\usepackage{amsmath}
\usepackage{amssymb}
\usepackage{mathtools}
\usepackage{amsthm}
\usepackage{multirow}
\usepackage{makecell}
\usepackage{tabularx}
\usepackage{subcaption}
\usepackage{wrapfig}
\usepackage{graphicx}
\usepackage[table]{xcolor}
\usepackage{pifont}
\usepackage{fvextra}
\usepackage{xspace}

\usepackage[utf8]{inputenc} 
\usepackage[T1]{fontenc}    
\usepackage{url}            
\usepackage{booktabs}       
\usepackage{amsfonts}       
\usepackage{nicefrac}       
\usepackage{microtype}      

\usepackage{color, soul}
\usepackage[most,breakable]{tcolorbox}
\usepackage{textgreek}

\usepackage[immediate]{silence}
\usepackage[labelfont=bf]{caption}
\usepackage{enumitem}
\usepackage{sectsty}
\usepackage{pgfplots}
\pgfplotsset{compat=1.18}

\usepackage{xcolor} 

\usepackage{geometry}

\usepackage{fancyhdr}

\makeatletter
\AtBeginDocument{%
  \geometry{headheight=20pt}
  \setlength{\headheight}{20pt}
  \pagestyle{fancy}
}
\makeatother

\renewcommand{\headrulewidth}{1pt}
\makeatletter
\def\headrule{{\if@fancyplain\let\headrulewidth\plainheadrulewidth\fi
\hrule\@height\headrulewidth\@width\textwidth \vskip-\headrulewidth}}
\makeatother

\definecolor{BMBDarkBlue}{HTML}{315EFE}
\definecolor{BMBLightBlue}{HTML}{00D3ED}
\definecolor{colhighlight}{HTML}{3B7BFF}
\definecolor{colhighlightlight}{HTML}{C8D6FF}
\definecolor{sectionbg}{HTML}{F7F7F7}

\usepackage[colorlinks, linkcolor=BMBDarkBlue, anchorcolor=BMBDarkBlue, citecolor=BMBDarkBlue, urlcolor=BMBDarkBlue]{hyperref}
\usepackage{cleveref}
\usepackage{threeparttable}

\sectionfont{\color{BMBDarkBlue}\fontfamily{zi4}\selectfont}
\subsectionfont{\color{BMBDarkBlue}\fontfamily{zi4}\selectfont}
\subsubsectionfont{\color{BMBDarkBlue}\fontfamily{zi4}\selectfont}

\newtcolorbox{mytheorem}{
  colback=gray!5,       
  colframe=gray!80,     
  boxrule=0.5pt,        
  arc=4pt,              
  left=4pt,             
  right=4pt,            
  top=4pt,              
  bottom=4pt,           
}

\newcommand{\fancyheadname}{\textit{\textbf{\modelname{}}}}

\title{\modelname{}: Realizing Long-Horizon Deep Exploration for Edge-Scale Agents}

\author{%
\textbf{\large{AgentCPM Team}}
}

\def\huggingface{\raisebox{-1.5pt}{\includegraphics[height=1.05em]{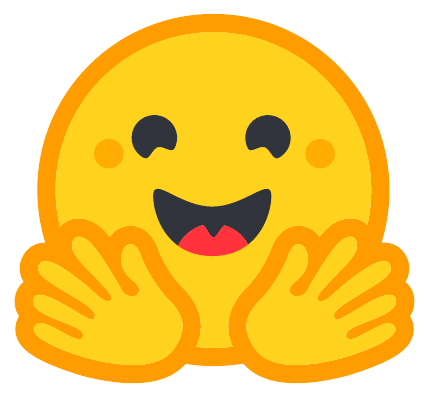}}}
\def\github{\raisebox{-1.5pt}{\includegraphics[height=1.05em]{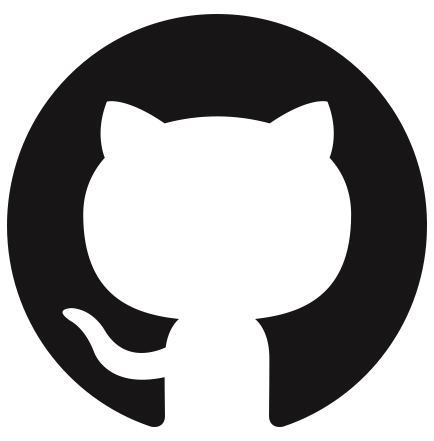}}}

\newcommand{\modelname}[0]{AgentCPM-Explore}

\usepackage{algorithm}
\usepackage{algpseudocode}
\usepackage{setspace}

\makeatletter
\renewcommand{\ALG@beginalgorithmic}{\small}
\makeatother



\begin{document}

\maketitle
\thispagestyle{fancy} 


%
\begin{abstract}
While Large Language Model (LLM)-based agents have shown remarkable potential for solving complex tasks, existing systems remain heavily reliant on large-scale models, leaving the capabilities of edge-scale models largely underexplored. In this paper, we present the first systematic study on training agentic models at the 4B-parameter scale. We identify three primary bottlenecks hindering the performance of edge-scale models: catastrophic forgetting during Supervised Fine-Tuning (SFT), sensitivity to reward signal noise during Reinforcement Learning (RL), and reasoning degradation caused by redundant information in long-context scenarios. To address the issues, we propose AgentCPM-Explore, a compact 4B agent model with high knowledge density and strong exploration capability. We introduce a holistic training framework featuring parameter-space model fusion, reward signal denoising, and contextual information refinement. Through deep exploration, AgentCPM-Explore achieves state-of-the-art (SOTA) performance among 4B-class models, matches or surpasses 8B-class SOTA models on four benchmarks, and even outperforms larger-scale models such as Claude-4.5-Sonnet or DeepSeek-v3.2 in five benchmarks. Notably, AgentCPM-Explore achieves 97.09\% accuracy on GAIA text-based tasks under pass@64. These results provide compelling evidence that the bottleneck for edge-scale models is not their inherent capability ceiling, but rather their inference stability. Based on our well-established training framework, AgentCPM-Explore effectively unlocks the significant, yet previously underestimated, potential of edge-scale models.

\end{abstract}

\begin{figure}[ht!]
    \centering
    \resizebox{0.99\linewidth}{!}{
        \includegraphics{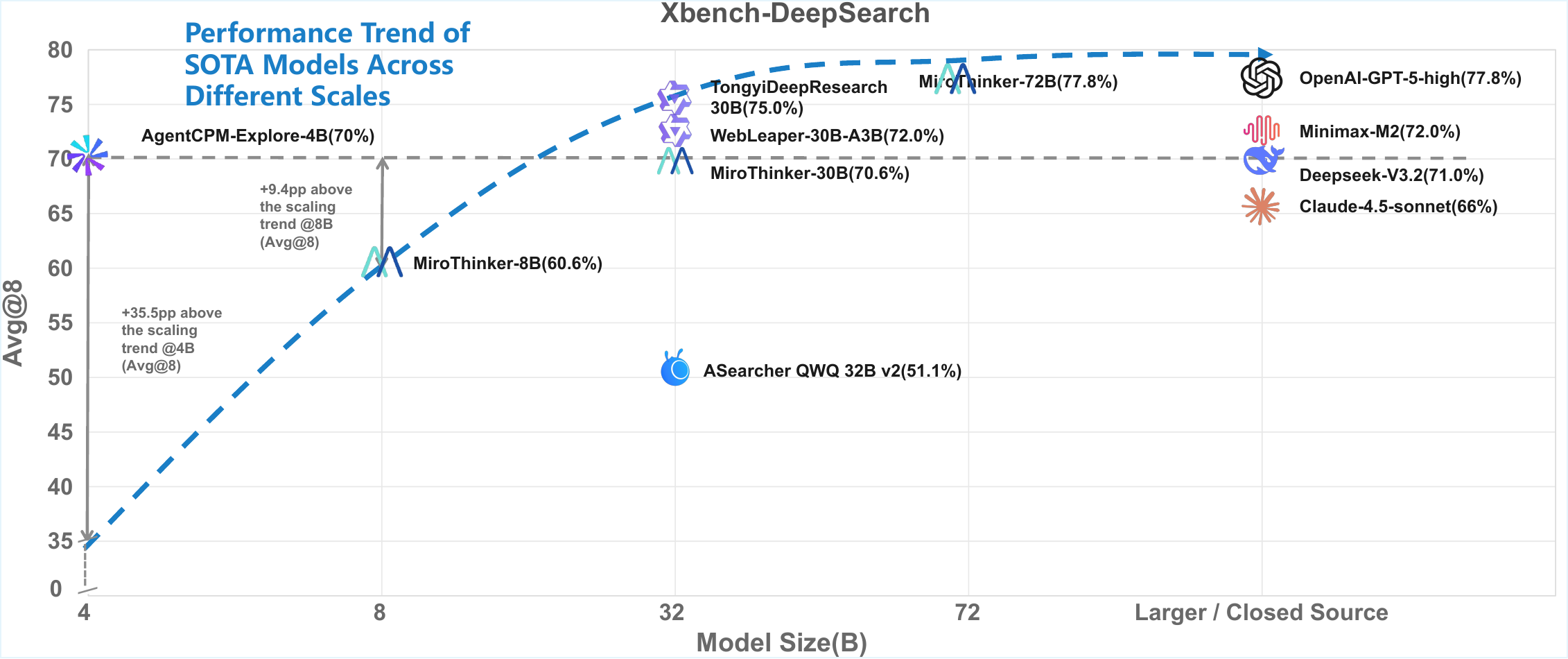}
    }
    \label{fig:cover}
\end{figure}


\newpage
{
  \hypersetup{linkcolor=RoyalBlue, linktoc=page}
  \tableofcontents
}

\newpage

\section{Introduction}

The proliferation of intelligent devices, coupled with growing user demands for privacy protection and real-time responsiveness, has positioned edge intelligence as a pivotal direction for deploying large language model (LLM) agents. This paradigm holds considerable promise across diverse application scenarios, including personal assistants, smart home systems, and mobile productivity tools. However, edge devices are inherently constrained by limited computational resources and power budgets, rendering them incapable of supporting the large-scale models (typically exceeding 8B parameters) upon which contemporary agent systems predominantly rely~\cite{tian2025clone,wang2025empowering,li2025pushing,liu2024edgellms,alizadeh2024llm}. Consequently, a pressing research challenge emerges: how to fully elicit advanced reasoning capabilities and complex task execution within a constrained parameter space.

This work represents the first systematic investigation into training agent models at the 4B parameter scale, with the aim of exploring the capability boundaries of edge-scale agent models (i.e., small agent models) and unlocking their long-horizon deep exploration capability. Compared to their larger counterparts, small agent models possess limited parameter capacity that struggles both to acquire specialized competencies while preserving general-purpose abilities, and to absorb training noise through redundant parameter space. These constraints render naive distillation or fine-tuning methods largely ineffective, giving rise to three fundamental challenges in small agent model training: 
(1) \textbf{Catastrophic forgetting during SFT}: When fitting specialized task-solving trajectories, small agent models tend to overwrite parameter space originally allocated to general capabilities, resulting in severe degradation of foundational abilities; 
(2) \textbf{Reward signal sensitivity during RL}: Small agent models exhibit limited tolerance to erroneous learning signals, namely reward noise, during reinforcement learning. Traditional agent RL paradigms lack precise credit assignment mechanisms, and sparse, noisy reward signals readily destabilize or even collapse the training process; 
(3) \textbf{Information contamination in long contexts during inference}: The redundant and noisy observations generated in real-world agent environments progressively erode the limited context modeling capacity of small agent models, leading to rapid deterioration in their decision quality.

To address these issues, we propose a systematic three-stage framework that leverages parameter-space model merging, reward signal denoising, and context information refinement.
First, we introduce a \textbf{parameter-space model-merging strategy} that fuses the weights of a general-purpose foundation model with those of a task-specific model. The strategy injects domain-specific skills while preserving general language understanding and reasoning capabilities, thereby mitigating overfitting. 
Second, we design a \textbf{reward signal denoising mechanism} that filters trajectories for which accurate credit assignment is infeasible due to environmental factors (e.g., network instability), formatting errors, or excessive interaction turns, thereby protecting the fragile learning dynamics of small agent models. 
Third, we propose a \textbf{context information refinement method} that operates on two fronts: incorporating search intent generation into the reinforcement learning optimization objective, and enhancing the information-extraction capability of summarization models through teacher model distillation. Together, these techniques improve the focus of environmental feedback summaries, achieving high-fidelity compression before information enters the context of the 4B model.

Experimental results demonstrate that AgentCPM-Explore achieves state-of-the-art performance among 4B-scale models across all eight evaluation benchmarks, matches or surpasses 8B-scale SOTA models on four of these benchmarks, and outperforms larger models (e.g., Claude-4.5-Sonnet, Kimi-Researcher, and DeepSeek-V3.2) as well as certain open-source models at the 30B parameter scale on four benchmarks. Notably, when afforded additional inference attempts, as exemplified by pass@64 results on the GAIA benchmark, AgentCPM-Explore successfully solves up to 97.09\% of text-based GAIA tasks. This finding provides compelling evidence that, built upon an appropriate training framework, edge-scale models possess the latent potential to address the vast majority of complex problems.

The remainder of this paper is organized as follows: Section 2 presents our methodology in detail, including the three core techniques; Section 3 reports the main experimental results and discusses the implications of this research; and Section 4 concludes the paper with directions for future work.




\section{Methodology}
The construction of LLM-driven agents typically follows a two-stage paradigm that combines supervised fine-tuning (SFT) with reinforcement learning (RL). During the SFT stage, high-quality teacher trajectories endow the model with initial capabilities for target tasks, while the RL stage further unlocks its exploration and learning potential through environmental interactions and reward feedback. Through extensive experiments, we identify three core challenges in training agent models at the 4B parameter scale: overfitting, reward signal noise, and context contamination. To address these challenges, this paper proposes three corresponding solutions: parameter-space model merging, reward signal denoising, and context information refinement. An overview of the proposed framework is illustrated in Figure~\ref{fig:pipeline}:

\begin{figure}[ht!]
    \centering
    \includegraphics[width=\linewidth]{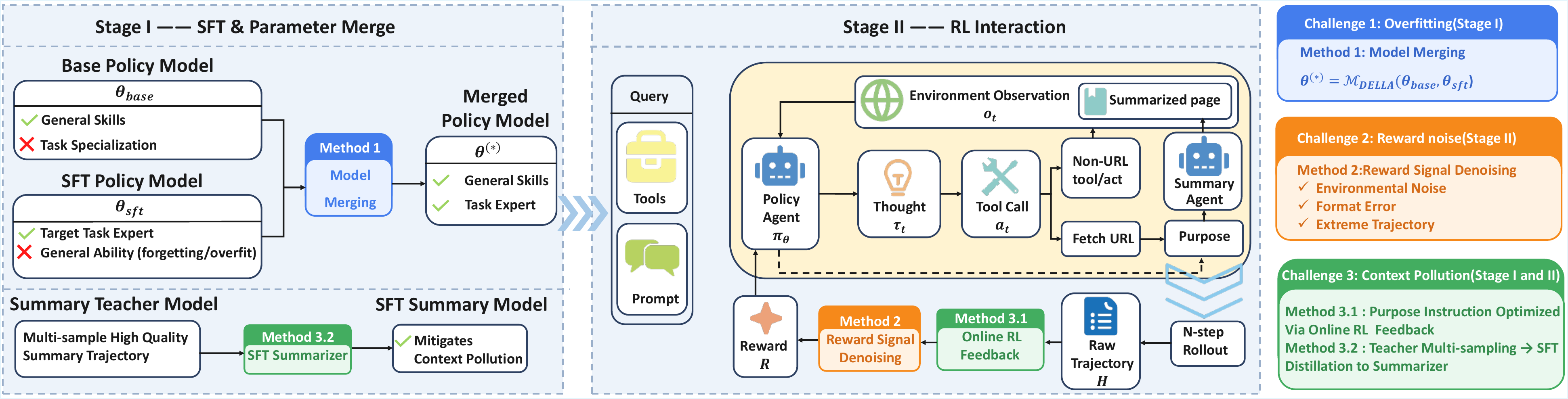}
    \caption{Overall Training Framework of AgentCPM-Explore.}
    \label{fig:pipeline}
\end{figure}

\subsection{Problem Formulation}

Agent tasks are typically formulated as Partially Observable Markov Decision Processes (POMDPs).
For deep research tasks, given a query $q$, the LLM-driven agent $\pi_{\theta}$ receives the current
context $C_t$ at time step $t$ and generates the thought $\tau_t$ (optional) along with an
action $a_t$:
\begin{equation}
\tau_t, a_t \sim \pi_{\theta}(\cdot \mid C_t), \quad a_t \in \mathcal{A},
\end{equation}
where $\mathcal{A}$ denotes the action space of the agent, which in this work corresponds to the set
of all available tools. Whether the thought $\tau_t$ is produced depends on the pretrained
model underlying the agent; since AgentCPM-Explore is built upon a reasoning model, it includes
$\tau_t$. The model continues to generate an interaction trajectory
\begin{equation}
H = \{(\tau_1, a_1, o_1), \ldots, (\tau_T, a_T, o_T)\}
\end{equation}
until yielding a final answer $a^{\mathrm{end}}$, where $\tau_t$, $a_t$, and $o_t$
denote the reasoning, action, and observation at step $t$, respectively.

The optimization objective for the LLM-driven agent is to identify a parameter configuration
$\theta^{\star}$ that maximizes performance on a test set comprising $N$ samples:
\begin{equation}
\theta^{\star} = \arg\max_{\theta}\; \frac{1}{N}\sum_{i=1}^{N}
\mathbf{1}\!\big[a^{\mathrm{end}}_i = a^{\mathrm{gt}}_i\big],
\end{equation}
where $a^{\mathrm{gt}}$ represents the ground-truth answer. However, directly optimizing this objective with a small agent model at the 4B parameter scale encounters three structural difficulties:
(1) the risk of overfitting in the initial capabilities of $\pi_{\theta}$ (where $\theta_{\mathrm{sft}}$ deviates from the general distribution); (2) the high signal-to-noise ratio requirement for the reward function $R(H)$ (e.g., false negatives induced by environmental noise); and (3) the dilution of information density in the context $C_t$ (irrelevant noise within observations $o_t$). We propose targeted solutions to address each of these three difficulties.

\subsection{Parameter-Space Model Merging}
During the SFT stage, we observe a critical phenomenon: models obtained by directly fine-tuning on target scenario data, while demonstrating improved performance on target tasks, exhibit more pronounced degradation in general capabilities (such as long-context comprehension and complex instruction following) than the base model. This degradation, in turn, impedes more substantial improvements and generalization on the target tasks themselves. This phenomenon is particularly pronounced in small models at the 4B parameter scale, manifesting as severe overfitting and catastrophic forgetting. Meanwhile, the base model retains general capabilities but lacks specialized execution abilities for the target scenario. This observation suggests that the two models possess complementary capability dimensions that the other lacks.

Building on this insight, we adopt a parameter-space model-merging strategy to achieve capability complementarity between the base and SFT models, enabling the merged model to inherit both the general comprehension abilities of the base model and the specialized execution capabilities of the fine-tuned model. General capabilities support the model in accurately understanding environmental feedback and long-context information, while specialized capabilities enable accurate decision-making in target scenarios. Consequently, the merged model, serving as the policy model for subsequent RL training, benefits from a higher-capability baseline, a broader exploration space, and an elevated optimization ceiling. Specifically, we employ the DELLA model merging algorithm~\cite{deep2024dellamerging}. Let $\theta_{\mathrm{base}}$ denote the
base model parameters, $\theta_{\mathrm{sft}}$ the fine-tuned model parameters, and
\begin{equation}
\delta^{(i)} = \theta_{\mathrm{base}}^{(i)} - \theta_{\mathrm{sft}}^{(i)}.
\end{equation}
The merged model parameters are computed as:
\begin{equation}
\theta^{(*)}_k = \theta^{(0)}_k + \lambda\; \frac{\sum_{i=1}^{2} w_i\; \mathbf{1}\!\left[\operatorname{sgn}\!\Big(\tilde\delta^{(i)}_k\Big)=\operatorname{sgn}\!\Big(\tilde\delta^{(1)}_k+\tilde\delta^{(2)}_k\Big)\right]\;\tilde\delta^{(i)}_k}{\sum_{i=1}^{2} w_i\;\mathbf{1}\!\left[\operatorname{sgn}\!\Big(\tilde\delta^{(i)}_k\Big)=\operatorname{sgn}\!\Big(\tilde\delta^{(1)}_k+\tilde\delta^{(2)}_k\Big)\right]+\varepsilon}\,,
\end{equation}
where
\begin{equation}
\tilde\delta^{(i)}_k = \frac{(1-z^{(i)}_k)\;\delta^{(i)}_k}{1-p^{(i)}_k},\quad z^{(i)}_k \sim \operatorname{Bernoulli}\!\big(p^{(i)}_k\big).
\end{equation}
Here, $p^{(i)}_k$ is assigned by MagPrune according to the magnitude rank of $\lvert \delta^{(i)}_k \rvert$, implementing a ``lower magnitude, higher drop probability'' scheme; the factor $\frac{1}{1-p}$ serves as the rescaling term to compensate for stochastic dropping. We set $\lambda = 0.9$ and $w_i = 1$, with no pruning dropout applied.
The hyperparameter $\lambda$ controls the ``displacement magnitude'' of the merged model relative to the base (global scaling/step size), determining the trade-off between preserving general capabilities and incorporating target-domain modifications. The weights $w_i$ govern the relative contributions of multiple source deltas $\delta$ (the mixing ratio in the weighted average). When all $w_i$ are equal, this degenerates to simple averaging.

\subsection{Reward Signal Denoising Mechanism}
During the reinforcement learning stage, accurate credit assignment constitutes the cornerstone of policy optimization. For agent tasks, the final reward $R$ of a trajectory $H$ is often sparse and highly noisy. This poses particular challenges for 4B-scale small agent models, whose parameter space is highly sensitive to gradient errors. Frequent false negative signals, where the model reasons correctly but fails due to environmental factors, formatting errors, or occasional missteps within extremely long trajectories, can severely distort the policy's value estimation, erroneously penalizing correct steps or exploratory behaviors, ultimately leading to training collapse.

To address this problem, we propose the principle of \emph{``filtering over attribution.''} Rather than designing sophisticated algorithms to extract signals from noise, we intercept and eliminate contaminated trajectories before gradient backpropagation. Based on the nature of error sources, we construct a three-tier filtering mechanism:
(1) environmental noise filtering,
(2) format error filtering, and
(3) extreme trajectory filtering.

Specifically, in the agent RL setting, given a task $q$, the model generates an interaction
trajectory $H=\{(\tau_1,a_1,o_1),\ldots,(\tau_T,a_T,o_T)\}$, where $\tau_t$, $a_t$, and $o_t$ denote the reasoning, action, and observation at step $t$, respectively. The environment returns a final reward $R(q,H)$. We examine each trajectory to determine whether the reward signal can be accurately attributed, identifying three categories prone to erroneous attribution. The first involves penalties caused by environmental factors, where the error lies not with the model but with tool or network quality issues. The second involves penalties for local parsing errors, such as formatting mistakes, which are concentrated in specific steps and should not contaminate the reasoning components of the entire trajectory. The third involves rewards or penalties following extremely short or long problem-solving processes, where overly short successful trajectories offer limited learning value, while overly long failed trajectories make it difficult to identify the error source.

\paragraph{Environmental Noise Filtering.}
For task failures caused by environmental instability, we identify failures attributable to environmental factors rather than model decisions by analyzing features such as tool return status codes, response time distributions, and error message patterns. Specifically, tool response times exceeding a preset threshold $T_{\mathrm{timeout}}$ are classified as timeout events; return status codes within the server error range (e.g., 500, 502, 503, 504) are classified as API fluctuations; and anomalous shifts in response time distributions are classified as network latency issues. For failed trajectories attributed to environmental noise, the system either removes them from the current training batch or sets their reward signals to neutral, preventing negative signals introduced by environmental uncertainty from contaminating the model policy.

\paragraph{Format Error Filtering.}
Some task failures, while originating from the model itself, bear limited relevance to the core reasoning capabilities and instead reflect the degree of mastery over output format specifications or degradation in long-context abilities. Such errors include JSON parsing errors (e.g., the model omits brackets \texttt{[]} required to enclose parameters), non-standard parameter formatting (e.g., special characters like \texttt{<answer>} or \texttt{<tool\_call>} missing half-width delimiters), response structures not conforming to requirements (e.g., the content within the \texttt{<answer>} special token is not a strict final answer but rather a report), and cases where the model falls into repetitive generation after exceeding a certain length until truncation is triggered, resulting in incorrect answers. Since such errors do not accurately reflect the quality of the model's reasoning, we do not backpropagate the outcome reward through the entire trajectory for these failures during actual training. Instead, we mask these trajectories to exclude them from gradient updates.

\paragraph{Extreme Trajectory Filtering.}
When trajectory length falls at either extreme, the attribution quality of reward signals degrades. For extremely short successful trajectories, they typically correspond to overly simple tasks for which the model already possesses sufficient capability, yielding minimal marginal benefit from continued learning; moreover, such trajectories enter the gradient update process more rapidly due to the shorter rollout steps, potentially causing the model to overfit to simple patterns. Accordingly, we set a trajectory length or interaction turn threshold $L_{\mathrm{min}}$ and exclude successful trajectories below this threshold from the core training set. For extremely long failed trajectories, when the number of interaction turns $|H|>L_{\mathrm{max}}$, the root cause of failure may be distributed anywhere along the trajectory. Backpropagating negative rewards to all steps would severely distort gradient estimation; hence, we directly discard such samples.

\subsection{Context Information Refinement}
Current mainstream approaches to context information refinement predominantly employ multi-model collaboration schemes~\cite{team2025tongyi,team2025mirothinker}, in which an independent summarization model is deployed to await tool call commands from the policy model for webpage URL content retrieval. Upon acquiring URL information, the summarization model automatically interprets the policy model's webpage-viewing intent to eliminate substantial redundancy from the returned URL content, returning information relevant to the summarization model's intent and enabling it to focus on extracting and analyzing key information. However, this approach is primarily constrained by two bottlenecks: (1) Intent communication bottleneck: whether the policy model can accurately express the information requiring attention when accessing specific URLs, and optimize corresponding instruction formulations according to the capability characteristics of the summarization model; (2) Information extraction bottleneck: whether the summarization model can accurately extract target information based on the policy model's requirements and provide faithful feedback. If these bottlenecks are neglected, the entire agent system will reach the ceiling of information transmission, and continued reinforcement learning training may even lead to performance degradation.

AgentCPM-Explore proposes an end-to-end context collaborative refinement framework comprising two optimization loops. The first loop is RL-based intent alignment. For query instructions generated by the policy model, we incorporate them into the overall RL training process. Using task success or failure as the feedback signal, we optimize the end-to-end policy model's capability to generate instructions that are ``more readily comprehensible to the summarization model.'' The second loop is a distillation-based summarization enhancement. To address the limitations in the 4B summarization model's capabilities, we introduce a high-performance teacher model (e.g., DeepSeek-V3.2-thinking) to sample across multiple task categories and construct high-quality summarization data. Through SFT, we distill the teacher model's information-extraction capabilities into the 4B summarization model, ensuring high-quality summaries when processing long texts.

Notably, the prompt design for the summarization model warrants careful consideration. Specifically, whether to include the original question has a significant impact on performance. We observe that including the question typically yields performance improvements, stemming primarily from two sources: first, the question information renders the summarization model's objective more explicit; second, for certain advanced reasoning tasks, the summarization model may leverage its own reasoning capabilities to directly derive the answer, while the policy model tends to continuously invoke search tools and consequently fails to answer correctly. To avoid the confounding effects of the latter case and ensure that performance improvements can be more clearly and interpretably attributed, we do not provide the original question to the summarization model in our experiments.

\section{Experiment}

\subsection{Experimental Settings}
\textbf{Client and Server Interaction}. All tool services are deployed using Docker containerization, with server-side timeouts set to 120 seconds and client-side timeouts set to 180 seconds. Search operations support automatic retry (up to 3 attempts), and webpage content retrieval employs the Jina Reader API with token truncation (capped at 95K tokens). 

\textbf{SFT Settings}. Supervised fine-tuning experiments are conducted on the Qwen3-4B-Thinking-2507 model using 8×A800 GPUs. We adopt the FSDP2 training strategy, combined with Ulysses sequence parallelism (degree 4), supporting ultra-long contexts of up to 128K tokens. Training utilizes bfloat16 precision with gradient checkpointing and padding removal optimizations enabled. The per-device micro-batch size is set to 1, yielding a global batch size of 32. For optimization, the learning rate is set to $1.5 \times 10^{-5}$ with a warmup ratio of 10\%, and training proceeds for 4 epochs.

\textbf{Model Merging Configuration}. We employ the DELLA merge method implemented via the mergekit library.\footnote{\url{https://github.com/arcee-ai/mergekit}} The hyperparameters are configured as follows: density = 1.0, weight = 1.0, and $\lambda = 0.9$.

\textbf{Evaluation Configuration}. During evaluation, we set temperature = 1, top-p = 1, presence penalty = 1, and max tokens = 16384.

\textbf{RL Experimental Settings}. Further training of the SFT model is conducted using 4 nodes of 8×A800 GPUs. During sampling, the presence penalty is disabled to ensure sampling from the policy's actual probability distribution. Training employs FSDP2 with sequence parallelism (degree 8), supporting contexts exceeding 128K tokens. Mixed-precision training balances performance and numerical precision. The per-device micro-batch size is set to 1, with a global batch size of $48 \times \text{number of machines}$. For optimization, the learning rate is set to $1.5 \times 10^{-6}$ with a warmup period of 10 steps.

\subsection{Main Results}

AgentCPM-Explore demonstrates exceptional parameter efficiency across mainstream agent evaluation benchmarks, including GAIA, HLE, BrowserComp, BrowserComp (ZH), WebWalker, FRAMES, XBench-DeepResearch, and Seal-0. Our model not only achieves state-of-the-art performance among models of comparable scale but also matches or surpasses SOTA models with twice the parameter count (8B-scale), such as MiroThinker-8B, while rivaling the performance of certain models exceeding 30B parameters and proprietary large models. The experimental results are presented in the following table:

\begin{table}[ht]
\begingroup
\renewcommand\arraystretch{1.1}
\setlength{\tabcolsep}{4pt}

\centering
\caption{Overall performance on 8 frequently adopted agent benchmarks.}
\label{table:overall-results}

\resizebox{\textwidth}{!}{%
\begin{tabular}{@{}lcccccccc@{}}
\toprule
\textbf{Benchmarks} &
\makecell{\textbf{GAIA}\\\textbf{(Text-Only)}} &
\makecell{\textbf{Browse}\\\textbf{Comp}} &
\makecell{\textbf{Browse}\\\textbf{Comp-ZH}} &
\makecell{\textbf{Humanity's}\\\textbf{Last Exam}} &
\textbf{FRAMES} &
\makecell{\textbf{WebWalker}\\\textbf{QA}} &
\textbf{SEAL-0} &
\makecell{\textbf{xbench}\\\textbf{DeepSearch}} \\
\midrule
\multicolumn{9}{l}{\textit{Closed-Source Models}} \\
\midrule
Claude-4.5-Sonnet~\cite{miromind2025mirothinker,anthropic2025claude45} & 71.2 & 19.6 & 40.8 & 24.5 & 85.0 & -- & 53.4 & 66.0 \\
Gemini Deep Research~\cite{gemini2025deepresearch} & -- & -- & -- & 26.9 & -- & -- & -- & -- \\
DeepSeek-V3.2~\cite{miromind2025mirothinker,deepseek2025v32} & 63.5 & 67.6 & 65.0 & 40.8 & 80.2 & -- & 38.5 & 71.0 \\
Minimax-M2~\cite{miromind2025mirothinker,minimax2025m2} & 75.7 & 44.0 & 48.5 & 31.8 & -- & -- & -- & 72.0 \\
OpenAI-GPT-5-high~\cite{miromind2025mirothinker,openai2025gpt5} & 76.4 & 54.9 & 65.0 & 35.2 & -- & -- & 51.4 & 77.8 \\
GLM-4.6~\cite{miromind2025mirothinker,zeng2025glm45,zai2025glm46} & 71.9 & 45.1 & 49.5 & 30.4 & -- & -- & -- & 70.0 \\
Kimi-Researcher~\cite{moonshot2025kimiresearcher} & -- & -- & -- & 26.9 & 78.8 & -- & 36.0 & 69.0 \\
Seed-1.8~\cite{bytedance2025seed18} & 87.4 & 67.6 & 81.3 & 40.9 & -- & -- & -- & -- \\
\midrule
\multicolumn{9}{l}{\textit{Open-Source Models}} \\
\midrule
MiroThinker 8B~\cite{miromind2025mirothinker} & 66.4 & 31.1 & 40.2 & 21.5 & 80.6 & 60.6 & 40.4 & 60.6 \\
Tongyi-DeepResearch 30B~\cite{team2025tongyi} & 70.9 & 43.4 & 46.7 & 32.9 & 90.6 & 72.2 & -- & 75.0 \\
ASearcher-QWQ-32B v2~\cite{gao2025beyond} & 58.7 & -- & -- & -- & 74.5 & -- & -- & 51.1 \\
IterResearch-30B-A3B~\cite{chen2025iterresearch} & 72.8 & 37.3 & 45.2 & 28.8 & 71.0 & -- & 39.6 & -- \\
WebSailor-V2-30B-A3B~\cite{li2025websailor} & 74.1 & 35.3 & 44.1 & 30.6 & -- & -- & -- & 73.7 \\
WebLeaper-30B-A3B~\cite{tao2025webleaper} & 73.2 & 38.8 & -- & -- & -- & -- & 48.6 & 72.0 \\
WebDancer-QwQ-32B~\cite{wu2025webdancer,li2025nested} & 51.5 & 3.8 & 18.0 & -- & -- & 47.9 & -- & 38.3 \\
\midrule
Merged-Model-4B & 60.1 & 21.9 & 26.8 & 17.4 & 80.3 & 65.4 & 35.9 & 69.9 \\
\rowcolor{colhighlightlight} AgentCPM-Explore-4B & 63.9 & 24.1 & 29.1 & 19.1 & 82.7 & 68.1 & 40.5 & 70.0 \\
\bottomrule
\end{tabular}%
}
\endgroup
\end{table}

Specifically, AgentCPM-Explore achieves optimal performance across all benchmarks among models of comparable scale (4B-level). More notably, it demonstrates remarkable cross-tier competitiveness. As a 4B model, AgentCPM-Explore surpasses models with 8× its parameter count (30–32B) on multiple benchmarks. For instance, it achieves 63.9\% on the GAIA subset, outperforming WebDancer (QWQ-32B) by 12.4 percentage points; on FRAMES, it reaches 82.7\%, surpassing ASearcher-32B by 8.2 percentage points and IterResearch-30B by 11.7 percentage points. When compared against substantially larger models such as DeepSeek-v3.2 and Kimi-Researcher, AgentCPM-Explore maintains its lead on multiple evaluations, including FRAMES and Seal-0, and exceeds the proprietary model Claude-4.5-Sonnet by 4.5 percentage points on BrowseComp. These results demonstrate that, built upon a well-established training framework and method, small models could bridge the parameter gap to compete with large models on agent tasks.

\subsection{Analysis of Context Information Refinement}
\begin{figure}[ht]
    \centering
    \includegraphics[width=0.9\linewidth]{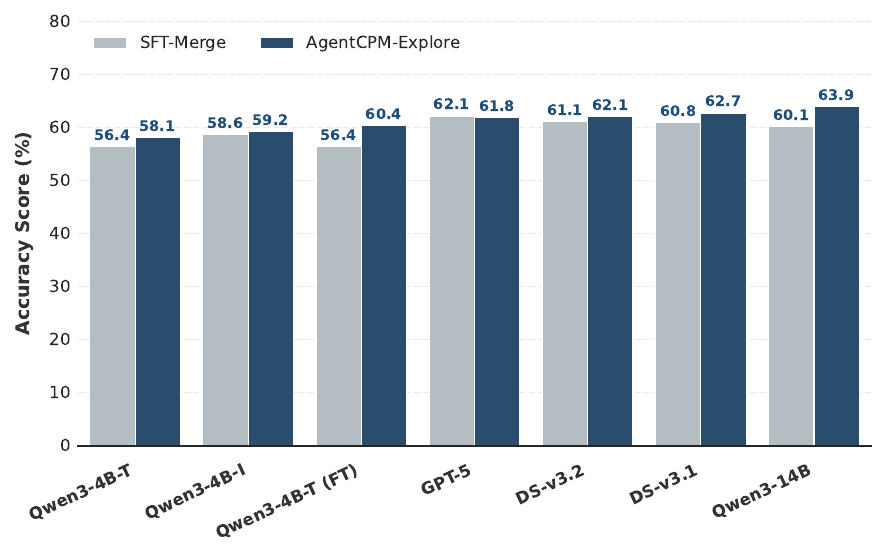}
    \caption{Overall performance of different summary models in the same agent system on the GAIA benchmark. Abbreviations are as follows. DS: DeepSeek; Qwen3-4B-I: Qwen3-4B-Instruct-2507; Qwen3-4B-T: Qwen3-4B-Thinking-2507; FT: Fine-tuned.
    }
    \label{fig:summary}
\end{figure}

To validate the necessity of the context information refinement method, we conduct summarization model substitution experiments. While keeping the agent model and all other system components completely unchanged, we substitute only the summarization model and observe the resulting changes in overall system performance. The experimental results are illustrated in Figure~\ref{fig:summary}. When employing summarization models of varying capabilities, the system shows significant performance differences on the GAIA benchmark. The experimental results demonstrate two key findings.


\textbf{The fidelity of information compression constitutes a critical bottleneck}. In multi-model collaborative agent systems, performance depends not only on the agent model's reasoning capability but also on the quality of information compression. With the agent model fixed, upgrading the summarizer, whether by swapping in a stronger model or through fine-tuning, consistently improves performance. Replacing Qwen3-14B with DeepSeek-V3.2 or GPT-5 raises GAIA accuracy from 60.1\% to 61.1\% and 62.1\%, respectively; SFT-based fine-tuning alone yields gains from 58.1\% to 60.4\%. When the summarizer has access to the original question, differences across summarizers are amplified to 10 percentage points (59\% → 69\%). After excluding cases where the summarizer directly solves the task, improvements in compression quality remain the primary source of gains. These results indicate that 4B-scale agent models possess sufficient reasoning potential for complex tasks, but are bottlenecked by lossy input compression.

\textbf{However, synergistic adaptation proves more effective than stronger components}. Through end-to-end RL, the agent learns to provide clear, well-structured purposes aligned with how the summarizer processes information. This trained agent, paired with a Qwen3-14B summarizer, achieves 63.9\%: outperforming even the GPT-5 summarizer setup (62.1\%). A well-adapted interface beats a more powerful but uncoordinated component. The path forward lies not in assembling stronger parts, but in enabling them to co-evolve.

\subsection{Analysis of Reward Denoising Mechanism}
\begin{figure}[ht]
    \centering
    \includegraphics[width=0.9\linewidth]{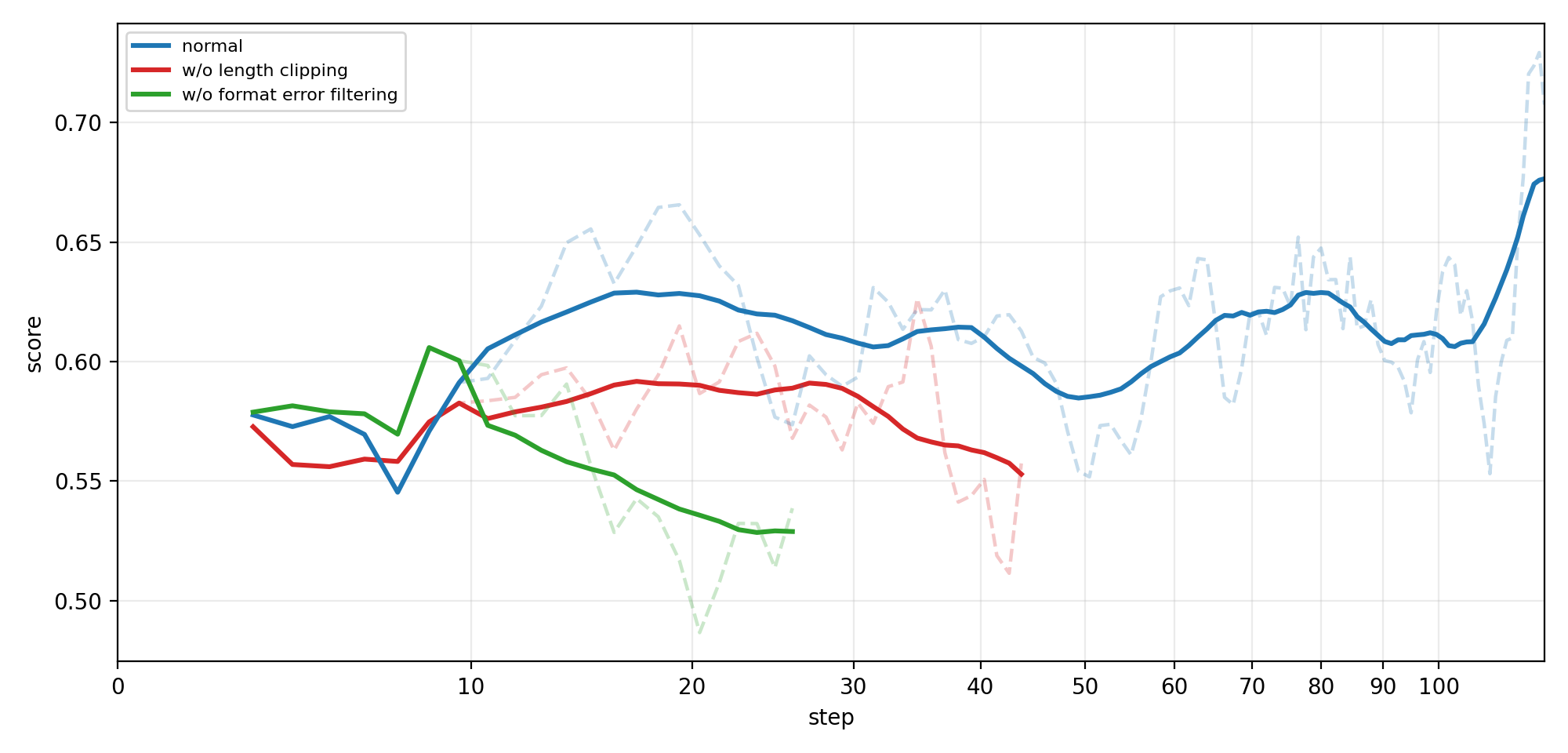}
    \caption{RL learning curves in different train settings.}
    \label{fig:reward-denoise}
\end{figure}

To assess the effectiveness of the reward signal refinement method, we visualize the policy learning curves before refinement in Figure~\ref{fig:reward-denoise}. To validate the effectiveness of the reward signal denoising mechanism, we conduct an ablation study. The blue curve represents AgentCPM-Explore with the complete three-tier filtering mechanism; the red curve represents a variant with extreme trajectory filtering removed; and the green curve represents a variant with format error filtering removed.

The experimental results reveal the extreme sensitivity of 4B-scale models to reward signal noise during the reinforcement learning stage. When any filtering mechanism is removed, model performance not only fails to improve but also degrades significantly. Combined with training dynamics analysis, we observe anomalous policy entropy collapse in the policy model under insufficiently filtered training environments. We attribute this to erroneous credit assignment induced by noisy signals: when the model incorrectly assigns penalties for tiny errors to correct reasoning steps, it gradually develops ``learned avoidance'' of effective exploratory behaviors, instead converging to conservative, myopic strategies or repetitive output patterns.

Furthermore, since environmental noise primarily manifests during the early stages of infrastructure development, it presents as a rapid collapse of training curves. Given that the experimental environment during this phase differs from the final training configuration of AgentCPM-Explore, the corresponding curves are not shown in the figure. However, this phenomenon further underscores the need for environmental noise filtering.

The practice of AgentCPM-Explore demonstrates that in long-horizon agent tasks where credit assignment is inherently imperfect, the "filtering over attribution" reward signal refinement strategy serves as a critical means for preserving the exploratory disposition of small models and maintaining training stability.

\subsection{Visualize the Capability Boundary of Edge-Scale Models}

\begin{figure}[ht]
    \centering
    \includegraphics[width=0.9\linewidth]{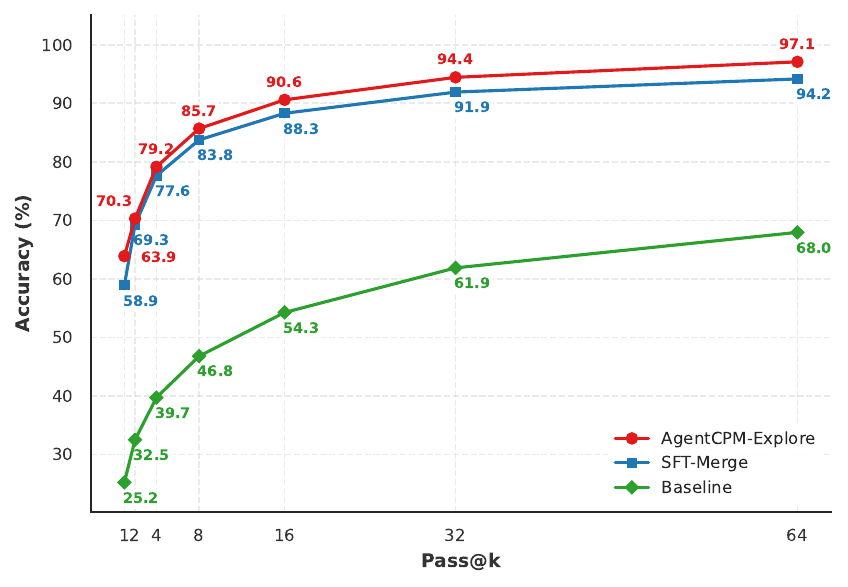}
    \caption{
        Pass@K performance comparison on the GAIA benchmark. The figure illustrates the capability boundary of the 4B model by comparing the ``SFT-Merge'' baseline with the final AgentCPM-Explore (RL after merging).
        The Qwen3-4B-thinking-2507 serves as the base model.
    }
    \label{fig:passk_merge_analysis}
\end{figure}

Having validated the information bottleneck hypothesis and broken through the system performance ceiling, a natural question arises: once bottleneck constraints are removed, where exactly lies the true capability boundary of edge-scale models (e.g., at the 4B-parameter level)? To address this question, we systematically map the Pass@K curves of the Qwen3-4B-Thinking-2507 model across different SFT and RL training stages, as illustrated in Figure~\ref{fig:passk_merge_analysis}. After validating the effectiveness of each system module, we attempt to answer a more fundamental question: when constraints from information bottlenecks and training framework algorithms are lifted, what is the actual capability ceiling of models at the 4B parameter scale?

Through systematic mapping of Pass@K curves across different SFT and RL training stages, we arrive at an encouraging finding: under Pass@64 sampling conditions, the SFT-trained model achieves an accuracy of 97.09\% on the GAIA benchmark. This result powerfully refutes the conventional assumption that ``edge-scale models lack sufficient logical reasoning capability'', revealing a critical insight: \textbf{edge models do not lack the latent capability to solve complex problems; their current performance bottleneck stems not from the model ``not knowing how'', but from the model ``not being stable''. In other words, correct problem-solving paths already exist within the model's policy space: the challenge lies in reliably eliciting them during single-shot inference}. The evolution pattern of Pass@K curves clearly delineates the influence of capability boundaries and performance stability. Specifically, as shown in the figure, RL training effectively "compresses" the superior performance achieved at high K values into lower K value ranges: the 94\% performance level attained at Pass@64 can be reproduced at Pass@32 after RL training. However, the leap from Pass@2 to Pass@1 poses significant challenges: performance variance increases markedly in this interval, and while RL training can raise the lower bound of Pass@1, it struggles to achieve stable high-level convergence.

We posit that the gap between Pass@2 and Pass@1 reveals a fundamental dilemma in small-model reinforcement learning: \textbf{how to endow small models with precise policy-value estimation capabilities}. Under high-K sampling, models can hedge against single-decision uncertainty through policy diversity. However, Pass@1 requires the model to directly hit the optimal policy in a single sample. This essentially demands that the model not only knows ``which policies are viable'' but also "which policy is most reliable." Small models face a dual challenge in this regard: on one hand, \textbf{the 4B parameter scale limits fine-grained modeling capacity for complex state spaces, making it difficult to distinguish between superficially similar but value-disparate policy branches}; on the other hand, \textbf{reward signals in agent tasks typically appear at the end of long trajectories, and the limited context modeling capacity of small models hinders accurate attribution of final rewards to early critical decision points}.

The above analysis shifts the focus of future research from ``how to enable models to solve problems'' to ``how to reduce inference variance''. To bridge the gap, potential optimization pathways include: (1) Explicit value modeling—training independent verifiers to assist models in policy pruning during inference, rather than merely imitating successful trajectories; (2) Metacognitive training—enabling models to assess their own confidence levels and proactively seek additional context when uncertainty is high, rather than acting blindly; (3) Hierarchical decision architectures—decoupling high-level policy selection from low-level action execution to reduce the difficulty of end-to-end credit assignment.

The practice of AgentCPM-Explore demonstrates that edge-scale models have the full potential to become general-purpose problem solvers. The key to unlocking this potential may lie not in making models better at exploration, but in making them more precise at selection.

\section{Conclusion}

This paper presents the first systematic exploration of the capability boundaries of agent models at the 4B parameter scale, proposing AgentCPM-Explore—an edge-scale agent model with high capability density. To address the three core challenges encountered in small model training—catastrophic forgetting, reward signal noise sensitivity, and long-context information contamination, we design a comprehensive three-stage training framework: parameter-space model merging to balance general capabilities with specialized skills, reward signal denoising mechanisms to safeguard the fragile reinforcement learning process, and context information refinement methods to achieve high-fidelity compression of environmental feedback. The experimental results not only validate the effectiveness of our proposed methods but also show that edge-scale models do not have a ceiling in their capability to solve complex problems. Through carefully designed training frameworks, small models can exhibit long-underestimated potential for solving complex tasks, providing empirical evidence for advancing edge intelligence.

Our future work will focus on three directions. First, refining information compression and memory mechanisms by exploring unified model architectures for multi-task joint optimization and designing effective historical information retrieval strategies. Second, optimizing the training pipeline by enhancing data diversity and readability, addressing overfitting issues in small-scale models, and concurrently improving evaluation benchmarks. Third, building a more robust training infrastructure to enable fault-tolerant training and precise checkpoint recovery for resumption. These efforts aim to advance edge-scale deep research agents toward greater capability, higher efficiency, and improved reliability.

\section{Contributions and Acknowledgments}
\modelname{} is the result of the collective efforts of all members of our team. 

\textbf{Project Lead}: Haotian Chen

\textbf{Contributors} (in alphabetical order): Haotian Chen, Xin Cong, Shengda Fan, Yuyang Fu, Ziqin Gong, Yaxi Lu, Yishan Li, Boye Niu, Chengjun Pan, Zijun Song, Huadong Wang, Yesai Wu, Yueying Wu, Zihao Xie, Yukun Yan, Zhong Zhang

\textbf{Project Supervisor}: Yankai Lin, Zhiyuan Liu, Maosong Sun




\newpage

\bibliographystyle{unsrtnat}
\bibliography{main,main_1}

\newpage

\appendix

\section{Infrastructure of AgentCPM-Explore}
To support future agent development, we also develop and release a comprehensive infrastructure for agent research and deployment. \textbf{AgentDock} provides a unified framework for tool management and sandboxed execution, enabling systematic configuration of tool invocation and environment isolation that would otherwise introduce substantial engineering overhead in agent experimentation. \textbf{AgentRL} implements a fully asynchronous reinforcement learning framework tailored to agent workloads, supporting efficient and scalable large-scale training. \textbf{AgentToLeaP} offers a one-click evaluation pipeline across major tool-use benchmarks, standardizing evaluation protocols and improving the reproducibility and comparability of experimental results. All components are released as open-source software.

To motivate the design choices behind this infrastructure, we next describe the system objectives and architectural considerations arising from Deep Research Agent workloads.

\subsection{Design Objectives and Architectural Overview}

As large language models continue to improve, LLM-based agents have evolved from single-turn question answering systems to agents capable of solving complex, long-horizon tasks. Deep Research Agents require autonomous planning in open environments, frequent invocation of heterogeneous external tools, and long-range exploration and reasoning to complete tasks such as information retrieval, knowledge synthesis, and report generation. Supporting the efficient execution and continuous improvement of such agents introduces significant infrastructure challenges.

From a runtime perspective, Deep Research Agents must interact with diverse tools, including search engines, web browsers, code interpreters, and document parsers. Existing frameworks often suffer from tight coupling between agents and tools, tool version incompatibilities, inconsistent interfaces, limited concurrency, and complex sandbox isolation requirements. Ensuring stability for long-running tasks while supporting dynamic tool extension and hot-swapping further increases system complexity. From a training perspective, improving agent capabilities relies on reinforcement learning with large-scale parallel rollouts. Conventional training pipelines struggle with long-horizon, high-concurrency workloads, leading to low GPU utilization, blocked training and inference pipelines, and sample bias introduced by asynchronous updates.

    

\begin{figure}[ht]
    \centering
    \includegraphics[width=1.0\linewidth]{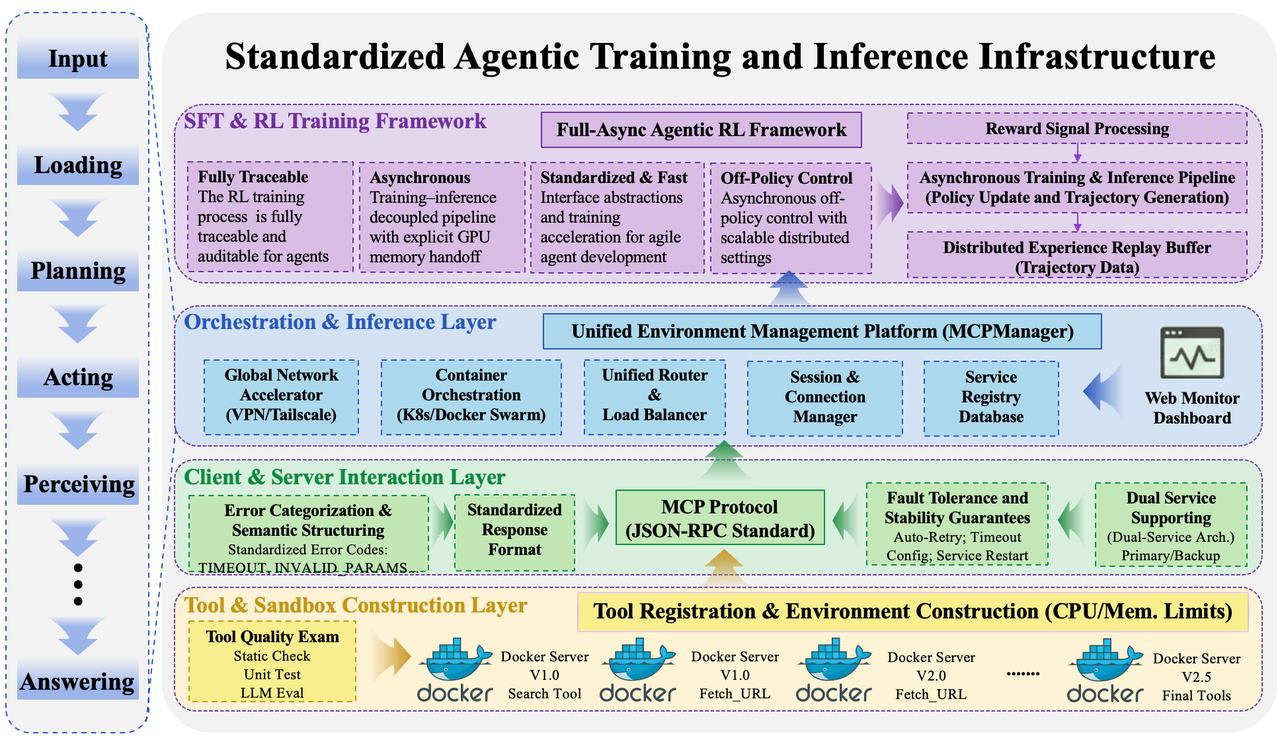}
    \caption{Standardized agentic training and inference infrastructure.}
    \label{fig:infra}
\end{figure}

To address these challenges, we design a full-stack infrastructure tailored for Deep Research Agents. As illustrated in Figure~\ref{fig:infra}, the system adopts a four-layer decoupled architecture comprising a Tool \& Sandbox Construction Layer, a Client \& Server Interaction Layer, an Orchestration \& Inference Layer, and an SFT \& Training Training Framework Layer. This design decouples infrastructure concerns from agent logic, enabling scalable deployment and independent evolution of system components.

\subsection{AgentDock}
\subsubsection{AgentDock: Tool \& SandBox Construction Layer}
The Tool \& SandBox Construction Layer addresses core challenges in agent tool ecosystems, including tool version incompatibility, heterogeneous sandbox requirements, and the lack of systematic quality validation. This layer constructs distributed tool execution environments based on Docker, ensuring strict isolation across tools while supporting composable and on-demand deployment.

For frequently invoked services such as search engines, web browsing, code execution, document parsing, and information summarization, we implement multi-version routing and load-balancing mechanisms to sustain stable high-concurrency execution, achieving over 100 QPS under stress-tested conditions. The layer integrates automated quality assurance procedures, including static validation, unit testing, and model-assisted evaluation. At the time of writing, the system supports 16 MCP service endpoints and integrates over one hundred tools with automated validation. For tools requiring external network access, global routing and acceleration are applied, reducing cross-region latency to the order of tens of milliseconds.

\paragraph{Tool Packaging and Registration.}

Tool packaging and registration constitute the entry point of the Tool \& Sandbox Construction Layer, addressing the standardized integration of heterogeneous open-source tools. All tools are encapsulated as MCP servers, with each tool deployed as an independent service within a dedicated Docker container.

\begin{itemize}[leftmargin=1.2em]
  \item \textbf{Packaging and metadata management.}
  The packaging process defines a structured capability schema for each tool, covering the tool name, functional description, parameter definitions, and return formats. Core execution logic is implemented within the containerized service, with explicit configuration of resource limits and runtime environments. Tool metadata and version information are centrally managed through TOML-based configuration files, enabling versioned registration, rapid lookup, and controlled evolution of tool interfaces.

  \item \textbf{Isolation and portability.}
  Containerized deployment ensures complete isolation across tools, allowing different runtime versions and dependency stacks to coexist without interference. Sandbox isolation guarantees that failures in individual tools do not affect other services or the host system, while standardized container images simplify tool deployment, migration, and replication.
\end{itemize}

\paragraph{Tool Quality Assurance and Filtering.}

Tool quality directly impacts agent execution reliability and the validity of training data. We establish a three-stage quality assurance pipeline consisting of static analysis, unit testing, and LLM-assisted evaluation.

\begin{itemize}[leftmargin=1.2em]

 \item  \textbf{Static Analysis.} Before deployment, automated static checks are performed to verify interface compliance and specification completeness. These checks include validation of input schemas, parameter type constraints, exception and timeout handling, and documentation completeness. Static analysis serves as the first quality gate, identifying specification-level issues without executing the tool.

 \item  \textbf{Unit Testing.} After passing static checks, standardized test cases are generated for each tool category to validate functional correctness. Representative queries are designed to assess semantic correctness of outputs, boundary condition handling, and robustness under invalid or extreme inputs. Response latency stability is also evaluated at this stage.
 
 \item For tool outputs that are difficult to assess using rule-based criteria, we introduce \textbf{LLM-assisted quality evaluation}. This includes reverse-query validation based on expected tool capabilities, semantic relevance scoring for retrieval-oriented tools, and consistency checks across repeated executions of identical queries. 
\end{itemize}

Tools are further filtered and prioritized based on task requirements. For a given task type, recommended tool sets are pre-selected and pre-warmed to reduce cold-start latency. Tool priorities are dynamically adjusted according to historical success rates.

\paragraph{High-Concurrency Deployment Support.}

In high-frequency invocation scenarios, single tool instances are insufficient to meet throughput requirements. We therefore adopt a multi-instance parallel deployment strategy.Each tool type can be deployed as a pool of container instances. For frequently invoked services such as search engines, web browsing, code execution, document parsing, and information summarization, sufficient instance replicas are provisioned to sustain concurrent workloads. Incoming requests are distributed across instances using round-robin or least-load scheduling strategies, preventing localized overload.

Elastic scaling is supported by dynamically adjusting the number of active instances based on request queue length and system load, balancing service capacity and resource utilization. Fault isolation mechanisms automatically remove failed instances from the pool without affecting other services. Through systematic stress testing, the current architecture sustains stable operation at a minimum throughput of 100 QPS for core services. For tools requiring external network access, global routing and network acceleration reduce cross-region latency to the order of tens of milliseconds.

\paragraph{Tool Capability Enhancement.}

Beyond baseline functionality, we enhance several core tools to better support agent workloads in Deep Research scenarios.

\begin{itemize}[leftmargin=1.2em]
\item  \textbf{Enhanced \texttt{read\_file} Tool.} The file reading tool is extended to support direct ingestion of documents referenced by URLs. The system automatically detects file types and parses online documents, including PDF, Word, PowerPoint, and image formats, converting them into unified textual representations. This significantly simplifies external knowledge acquisition for agents.

\item \textbf{Context Management.} To address context overflow in long-horizon research tasks, we implement an active context management mechanism. Token usage is continuously monitored, with intelligent truncation strategies that preserve critical information at the beginning and end of interaction histories while compressing intermediate content. A summarization module invokes a dedicated Summary Agent to generate concise summaries of historical interactions, and a rolling window mechanism retains full context for the most recent $N$ turns. This design enables effectively unbounded interaction lengths without being constrained by fixed context windows.

\item \textbf{Enhanced Code Executor.} The code execution tool is augmented with flexible environment configuration capabilities. Common data science libraries such as \texttt{numpy}, \texttt{pandas}, \texttt{matplotlib}, and \texttt{scikit-learn} are pre-installed, with documentation provided for agent reference. The executor supports runtime installation of additional dependencies and exposes command interfaces that allow agents to adjust execution environments dynamically, enabling a wide range of data analysis and programming tasks.
\end{itemize}

\subsubsection{AgentDock: Client \& Server Interaction Layer}
The Client \& Server Interaction Layer sits above the Tool \& SandBox Construction Layer and addresses tight coupling between agents and tools, as well as instability during tool invocation. This layer standardizes agent tool communication through the Model Context Protocol (MCP), establishes a unified client server communication mechanism across heterogeneous sandbox environments, and provides comprehensive fault-tolerance strategies for long-running tasks. These design choices enable stable support for training data collection, model optimization, and inference workloads over extended execution horizons.

\paragraph{Server Standardization}

Server standardization forms the foundation of the Client \& Server Interaction Layer, enabling agents to interact with arbitrary tools in a uniform manner.

\begin{itemize}[leftmargin=1.2em]
\item \textbf{MCP Adoption.}
We adopt the Model Context Protocol (MCP) as the standard communication protocol between agents and tools. All tools follow a unified JSON-RPC communication interface, allowing agents to invoke tools without knowledge of internal implementations. Tool capabilities are abstracted into structured interface specifications, which agents can parse to understand tool functionality, parameter requirements, and return formats.

\item \textbf{Unified Response Format.}
To simplify model parsing and the design of reinforcement learning rewards, we define a unified structured response format. Successful responses include a \texttt{status} field (set to \texttt{success}), a \texttt{content} field containing the returned result, and a \texttt{metadata} field providing auxiliary information. Error responses include a \texttt{status} field (set to \texttt{error}), an \texttt{error} field indicating the error type, a \texttt{detail} field providing diagnostic information, and an \texttt{error\_code} field returning a standardized error code. This unified format eliminates tool-specific parsing logic and provides a structured foundation for constructing reward signals. The Server layer explicitly encapsulates tool versions, allowing multiple versions of the same tool to coexist. Agents may specify tool versions on demand, preventing unintended disruptions to ongoing tasks caused by tool upgrades.
\end{itemize}

\paragraph{Error Classification and Semanticization}

Clear error semantics are essential for autonomous error recovery in agent systems. We design a multi-level error classification scheme that assigns explicit semantics and handling strategies to tool failures.

\begin{itemize}[leftmargin=1.2em]
  \item \textbf{Standardized error codes.}
  Tool failures are mapped to fine-grained categories with predefined recovery actions, including \texttt{TIMEOUT}, \texttt{INVALID\_PARAMS}, \texttt{SERVICE\_UNAVAILABLE}, \texttt{RATE\_LIMITED}, and \texttt{CONTENT\_TRUNCATED}.

  \item \textbf{Semantic normalization.}
  Error messages are normalized through a unified pipeline that removes irrelevant implementation details, attaches recovery hints, and provides structured signals for both runtime handling and reward shaping during training.
\end{itemize}

\paragraph{Fault Tolerance and Stability Guarantees}

To ensure reliable execution under long-running workloads, the Client \& Server Layer implements multi-level fault-tolerance mechanisms.

\begin{itemize}[leftmargin=1.2em]
  \item \textbf{Layered timeouts and retries.}
  A hierarchical timeout design spans client, server, MCP, and tool request layers, combined with bounded automatic retries using exponential backoff to mitigate transient failures.

  \item \textbf{Service-level redundancy.}
  Critical tools adopt a primary--secondary service architecture with multi-provider support and graceful degradation to preserve task completeness. Unresponsive MCP servers are automatically restarted.

  \item \textbf{Fallback routing.}
  When all instances of a tool become unavailable, requests are redirected to functionally similar fallback tools to prevent task termination.
\end{itemize}

\paragraph{Client--Server Communication Mechanism}

The Client \& Server Layer establishes a unified communication abstraction that shields agents from sandbox heterogeneity.

\begin{itemize}[leftmargin=1.2em]
  \item \textbf{Unified client abstraction.}
  Agents invoke tools exclusively through a single MCP client interface, which handles request serialization, transmission, response handling, and deserialization, providing consistent invocation semantics across all tools.

  \item \textbf{Sandbox adaptation and connection management.}
  An adapter-based design masks differences across sandbox environments (e.g., standard, GPU-enabled, or privileged containers), while persistent connections with health checks and automatic reconnection reduce communication overhead and improve reliability.
\end{itemize}

By combining protocol standardization, semantic error handling, fault tolerance, and unified communication, the Interaction Protocol Layer abstracts heterogeneous tool resources into stable and semantically transparent services. This abstraction provides a robust foundation for the Orchestration \& Inference Layer and supports long-horizon execution and continuous agent training.

\subsubsection{AgentDock: Orchestration \& Inference Layer}
The Orchestration \& Inference Layer addresses challenges in dynamic tool management and stable access for agent systems. In real-world Deep Research Agent workloads, tool orchestration faces three primary difficulties: (i) severe fragmentation of tool ecosystems, where heterogeneous dependency stacks, frequent version iterations, and the need for parallel multi-version execution complicate integration and reuse; (ii) tight coupling between tool invocation and underlying infrastructure, requiring callers to manage low-level details such as IP addresses, port mappings, and container states, leading to brittle systems and high maintenance costs; and (iii) unstable cross-region network access, which is a dominant source of tool invocation failures in practice, particularly for overseas search engines and web services.

To address these challenges, we construct a unified container management platform, termed \textbf{MCPManager}. The platform provides container orchestration, unified routing, session management, and global network acceleration as backend services, allowing agents to invoke tools solely through abstract tool capabilities without awareness of execution environments, network conditions, or concurrency control mechanisms.
\paragraph{Platform Lifecycle Management}

To address tool ecosystem fragmentation and complex version management, MCPManager provides full-lifecycle support for tool containers, covering onboarding, version control, and operational maintenance. Custom MCP tools are packaged by encapsulating code and runtime environments into container images, which can be released either as standalone tools or as part of versioned tool collections. Multiple tool collection versions with customized configurations can be maintained concurrently. Tool onboarding is streamlined to a minimal configuration process that requires only image registration, port exposure, and resource limit specification (e.g., 4 CPU cores and 16\, GB memory), after which the platform automatically assigns stable access endpoints and registers metadata in a centralized database.

The platform supports concurrent execution of multiple tool versions, enabling comparative testing and gradual migration. New versions are integrated through automatic routing updates, while legacy versions are gracefully phased out via rolling updates without service interruption. Operational support includes one-click start and stop operations to reduce cold-start overhead, periodic health checks with automatic restarts, and a real-time web dashboard that visualizes CPU usage, memory consumption, and container status. Management interfaces are secured through key-based authentication, with SSH public keys and access permissions managed via user configuration profiles, enabling safe multi-user sharing while maintaining isolation across independently managed container instances.

\paragraph{Unified Routing and Dynamic Scheduling}

To decouple tool invocation from underlying infrastructure details, MCPManager implements a unified routing and dynamic scheduling mechanism. Tools are exposed via container-ID-based proxy URLs managed by the platform, allowing agents to invoke tools using only tool names and version identifiers, without awareness of IP addresses, port mappings, or container states. All tool containers are accessed through a single external port, significantly simplifying network configuration while preserving consistent invocation semantics.

Infrastructure changes such as container restarts, port remapping, or IP reassignment are fully transparent to callers. Newly launched tool versions are automatically registered and become immediately available, enabling seamless hot-plugging. At runtime, the routing layer performs request-level health checks, triggers automatic restarts upon failure detection, and maintains stable external URLs even when backend ports change. Load balancing, rate limiting, service discovery, and fault recovery are handled transparently by the platform, ensuring reliable and uninterrupted tool access under dynamic workloads.

\paragraph{Global Network Acceleration}

To mitigate instability in cross-region tool access, MCPManager incorporates a global network acceleration mechanism. All containers are connected through a unified virtual private network built on Tailscale, enabling end-to-end encrypted communication and stable access via fixed internal IP addresses. This design decouples service access from physical network changes, simplifying connectivity across distributed deployment environments.

Outbound traffic is dynamically routed based on query language and target services. Chinese-language queries are routed through domestic endpoints, while English-language queries are routed through overseas egress nodes (e.g., US or UK). Containers access a host-level proxy via \texttt{host.docker.internal}, allowing the proxy to dynamically select optimal outbound routes, including Jina-based routing for web crawling tasks. As a result, the global networking strategy improves cross-border access success rates from approximately 70\% to over 95\%, reduces average latency by 30--50\%, and sustains high-concurrency workloads without triggering upstream rate limits.

\subsection{AgentRL: SFT \& RL Training Framework Layer}

The Training Framework Layer provides a fully asynchronous infrastructure for large-scale reinforcement learning of agent systems. The framework adopts a database-centric design that completely decouples inference-time sampling from policy optimization. Inference workers continuously generate interaction trajectories and persist them to storage, while training workers consume trajectories in a streaming manner to perform parameter updates. This design supports multi-task parallel sampling, continuous policy updates, and unified modeling of long-horizon multi-round interactions.

\paragraph{Fully Recorded and Auditable Training Pipeline}

The training framework implements a fully recorded, auditable pipeline in which all sampling requests, model responses, environment feedback, evaluation outcomes, and training consumption events are persisted in a structured format, serving as a complete source of truth for the entire training process. This design enables trace-level inspection of historical trajectories, allowing users to examine agent decisions across multi-round interactions and precisely locate anomalous samples.

By retaining full execution records throughout asynchronous training, the framework substantially reduces debugging and analysis overhead. As a result, rapid policy iteration and system-level diagnosis remain feasible even under large-scale, high-throughput training workloads.

\paragraph{Asynchronous Pipeline with Explicit Memory Handover}

The training framework adopts a fully asynchronous pipeline that explicitly decouples inference-time sampling from policy optimization. Sampling is modeled as a pluggable stage within the training loop, with explicit weight synchronization and model--optimizer state transfer enabling controlled transitions between inference and training phases. This design avoids concurrent residency of training and inference workloads on the same devices.

To prevent implicit GPU memory contention, the framework introduces an explicit memory handover mechanism that transfers memory ownership across phases rather than sharing it implicitly. By converting hidden resource competition into explicit scheduling decisions, the system significantly improves GPU utilization and stabilizes throughput under fully asynchronous execution.

\paragraph{Agent-Centric API Abstractions and Training Acceleration}

Agent construction is abstracted into composable API-level interfaces that allow users to define observation processing, action generation, and context management logic in a service-like manner, and to directly integrate agents into the reinforcement learning pipeline. This abstraction substantially lowers the development and iteration cost for multi-round agents.

To accelerate training, the framework introduces an automatic prefix merging mechanism that centrally manages shared context prefixes across multi-round trajectories. For stepwise reasoning agents such as ReAct, repeated prefixes are merged and reused, compressing multiple trajectory forward passes into a single forward computation. This achieves training-time computation reuse, analogous to inference-time KV caching, and significantly reduces redundant computation.

\paragraph{Asynchronous Off-Policy Control and Distributed Scaling}

\begin{itemize}[leftmargin=1.2em]
  \item \textbf{Off-policy drift mitigation.}
  To address distributional drift introduced by full asynchrony, the framework applies trajectory freshness constraints and bounded importance weighting to stabilize optimization.

  \item \textbf{Distributed scalability.}
  The training layer supports FSDP2 as well as tensor, context, and pipeline parallelism (TP/CP/PP), enabling efficient scaling to large models and long-context agent training workloads while maintaining stable convergence under high-throughput execution.
\end{itemize}

\paragraph{System-Level Bottlenecks in Multi-Turn Agent RL}

In multi-turn agent tasks, the primary efficiency bottleneck of reinforcement learning often arises not from the policy optimization algorithm itself, but from system-level structural friction. Unlike single-step RL environments, agents operating in realistic settings often use external tools, interact with high-latency systems, and generate trajectories of variable length and quality. As a result, training processes are commonly stalled by sampling delays, GPU memory contention, or uncontrolled concurrency, leading to low GPU utilization and prolonged parameter update intervals.

\begin{figure}[ht]
    \centering
    \includegraphics[width=0.8\linewidth]{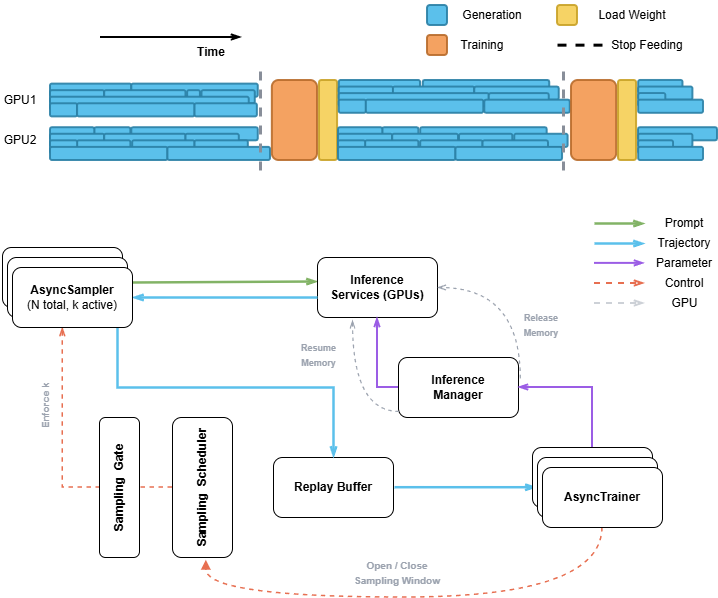}
    \caption{Overview of the Standardized Agentic Training and Inference Infrastructure.}
    \label{fig:arl}
\end{figure}

\paragraph{Fully Asynchronous AgentRL Architecture}
To address inefficient GPU utilization in agent RL, we design \textbf{AgentRL}, a fully asynchronous training framework that decouples sampling, inference, and training into independently operating subsystems via an external database. Multiple parallel \textit{AsyncSamplers} continuously issue generation requests to inference services, while completed trajectories are persistently stored. An \textit{AsyncTrainer} asynchronously consumes finalized trajectories from the database and performs policy updates. Crucially, sampling and training do not communicate through synchronous queues, preventing global stalls caused by blocking on either side. Figure~\ref{fig:arl} illustrates the overall architecture of AgentRL, including the asynchronous interaction between samplers, inference services, the replay buffer, and the training process, as well as the explicit coordination of GPU memory between inference and training phases.

An \textit{Inference Manager} explicitly coordinates model parameters and GPU memory usage between inference and training. During policy updates, inference requests are temporarily paused, and GPU memory is released for training. Once training completes and updated weights are loaded, the sampling window is reopened, and inference throughput is rapidly restored.

\paragraph{Concurrency Control and Explicit Resource Scheduling}

During sampling, AgentRL injects sufficient concurrent generation requests to keep inference GPUs saturated, preventing idle compute caused by slow environment interactions or long-horizon trajectory generation. Instead of unbounded concurrency, the Inference Manager dynamically regulates active sampling based on target concurrency and in-flight requests, ensuring high utilization while avoiding memory overflow or latency collapse.

During training phases, sampling is explicitly suspended, and inference memory is handed over to training workloads. After parameter updates, inference resumes immediately with refreshed weights, enabling rapid alternation between sampling and training without implicit resource contention.

\subsection{AgentToLeaP: Evaluation \& Validation Framework Layer}

The Evaluation Framework Layer provides a standardized infrastructure for the rigorous benchmarking of agentic systems. The framework adopts a unified execution protocol that decouples task definition from agent inference, normalizing heterogeneous evaluation datasets into a consistent interaction interface. By coordinating with the AgentDock layer for isolated tool execution and employing a multi-process orchestration engine, AgentToLeaP supports high-throughput, reproducible validation across complex, long-horizon tasks.

\paragraph{Standardized Protocol for Heterogeneous Benchmarks}

The framework implements a modular adapter mechanism to standardize input-output protocols across diverse agentic tasks. It unifies distinct task formats---ranging from general assistant queries (e.g., GAIA) to web browsing contexts (e.g., WebWalkerQA, BrowseComp) and long-context reasoning (e.g., HLE)—into a canonical evaluation interface. This abstraction allows researchers to validate a single agent configuration against multiple domains without developing ad-hoc execution scripts for each specific dataset.

\paragraph{High-Throughput Parallel Orchestration}

To address the computational latency inherent in multi-turn agent interactions, the framework incorporates a process-based parallel execution engine. The engine dynamically schedules evaluation tasks across available computational resources, maximizing inference throughput while maintaining strict environment isolation. This design significantly reduces the turnaround time for large-scale experiments compared to sequential execution baselines.

\paragraph{Structured Trace Logging and Persistence}

The framework establishes a comprehensive logging pipeline that persists complete interaction histories for analysis. All agent decisions, including Chain-of-Thought reasoning, tool invocations, and raw environment observations, are serialized into structured JSON formats (e.g., \texttt{dialog.json}, \texttt{trace.json}). This structured serialization ensures full reproducibility of experiments and enables researchers to perform granular, trace-level inspection of the agent's decision-making process.

\paragraph{Extensibility via Configuration}

The architecture follows a data-driven design pattern to facilitate rapid extension. New benchmarks can be integrated by simply adding datasets in the standardized JSONL format, removing the need for complex code registration. Furthermore, agent behaviors and reasoning strategies are managed through modular execution scripts, allowing users to customize agent baselines and interaction logic with minimal engineering overhead.

\section{Experimental Analysis of AgentRL}

\begin{figure}[ht]
    \centering
    \includegraphics[width=1.0\linewidth]{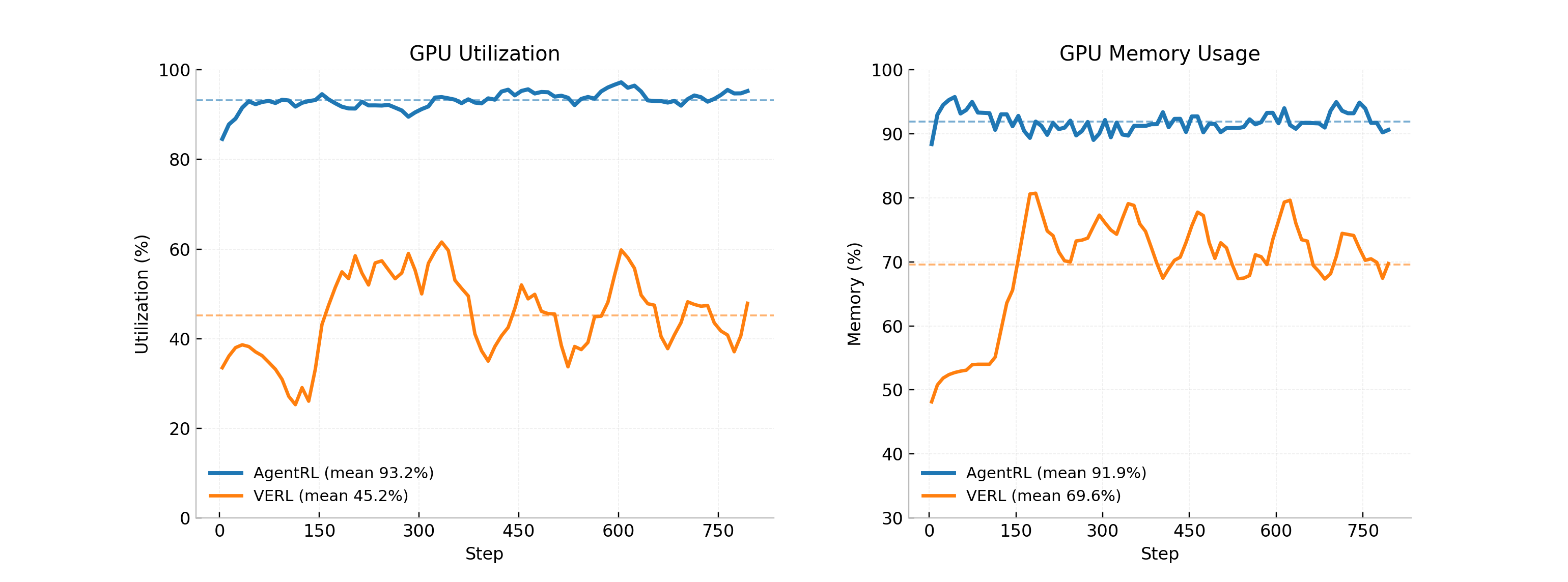}
    \caption{Comparison of GPU Utilization Between veRL and AgentRL.}
    \label{fig:arlcompar}
\end{figure}

We assess the effectiveness of the proposed system design by comparing GPU utilization with the widely used \textit{veRL} framework. As shown in Figure~\ref{fig:arlcompar}, AgentRL maintains consistently high GPU compute utilization (mean 93.2\%) together with stable memory occupancy (mean 91.9\%) throughout training, indicating that GPU resources remain continuously engaged over time. In comparison, \textit{veRL} exhibits more variable utilization patterns, with lower average compute and memory usage (45.2\% and 69.6\%, respectively), suggesting intermittent idle periods during training.

Notably, the observed efficiency differences do not stem from faster individual model computations. Rather, they arise from differences in the organization of system-level training. AgentRL adopts a fully asynchronous pipeline with explicit resource scheduling and memory handover between the sampling and training phases, thereby reducing idle time caused by synchronization and coordination overhead. This design enables AgentRL to sustain high utilization even under long-horizon, tool-intensive agent workloads. For clarity, GPU statistics are downsampled by averaging every 10 training steps and then smoothed with a moving average to highlight overall trends while reducing short-term noise.

\end{document}